\documentclass[10pt,twocolumn,letterpaper]{article}

\usepackage[pagenumbers]{iccv} 

\usepackage{multirow}
\usepackage{xcolor}
\usepackage{enumitem}

\usepackage[linesnumbered,ruled,vlined, noend]{algorithm2e}

\SetCommentSty{mycommfont}

\usepackage{caption}
\usepackage{xspace}

\newcommand{\sysname}{\textsc{CARS}\xspace}

\definecolor{iccvblue}{rgb}{0.21,0.49,0.74}
\usepackage[pagebackref,breaklinks,colorlinks,allcolors=iccvblue]{hyperref}

\def\paperID{3223} 
\def\confName{ICCV}
\def\confYear{2025}

\title{Making Every Frame Matter: Continuous Activity Recognition in Streaming Video via Adaptive Video Context Modeling}
\author{Hao Wu\\
Nanjing University
\and
Donglin Bai\\
Microsoft Research
\and
Shiqi Jiang\\
Microsoft Research
\and
Qianxi Zhang\\
Microsoft Research
\and
Yifan Yang\\
Microsoft Research
\and
Xin Ding\\
Microsoft Research
\and
Ting Cao\thanks{corresponding author}\\
Microsoft Research
\and
Yunxin Liu\\
AIR, Tsinghua University
\and
Fengyuan Xu\\
Nanjing University
}

\begin{document}
\maketitle

\begin{abstract}
Video activity recognition has become increasingly important in robots and embodied AI. 
Recognizing continuous video activities poses considerable challenges due to the fast expansion of streaming video, which contains multi-scale and untrimmed activities.
We introduce a novel system, \sysname, to overcome these issues through adaptive video context modeling. 
Adaptive video context modeling refers to selectively maintaining activity-related features in temporal and spatial dimensions.
\sysname has two key designs. 
The first is an activity spatial feature extraction by eliminating irrelevant visual features while maintaining recognition accuracy.
The second is an activity-aware state update introducing dynamic adaptability to better preserve the video context for multi-scale activity recognition.
Our \sysname runs at speeds $>$30 FPS on typical edge devices and outperforms all baselines by 1.2\% to 79.7\% in accuracy. 
Moreover, we explore applying \sysname to a large video model as a video encoder.
Experimental results show that our \sysname can result in a 0.46-point enhancement (on a 5-point scale) on the in-distribution video activity dataset, and an improvement ranging from 1.19\% to 4\% on zero-shot video activity datasets. 
\end{abstract}

\section{Introduction}

Video activity recognition~\cite{madan2024foundation,chen2024video,huang2018makes,zolfaghari2018eco,wu2021towards,tang2023video} is an important and long-standing application.
These tasks are widely used in advanced human-computer interactions (HCI), including robotics and augmented reality~\cite{duan2022survey,strafforello2023current,gadre2022continuous}. 
To enable advanced HCI, real-time recognition of streaming video on resource-constrained mobile or edge devices is crucial. 
Activity recognition relies heavily on video context, making it difficult to determine actions from a single frame. 
Therefore, \textit{a continuous video activity recognition system must maintain video context, allowing it to infer activities in new frames based on the accumulated context}.
However, achieving efficient continuous recognition in streaming video is challenging for the following reasons.


\begin{figure}[t]
  \centering
  \includegraphics[width=0.49\textwidth]{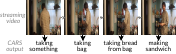}
  \caption{An example of continuous video activity recognition in streaming video. \sysname takes every frame as input and continuously recognizes human activities.} 
  \label{fig:scenario}
\end{figure}

\textbf{(i) Diverse Temporal Scales of Activities.} 
Continuous video activity recognition needs to capture the spatio-temporal features of activity on multiple temporal scales~\cite{zhao2018trn,shao2020finegym}. 
As shown in Figure~\ref{fig:scenario}, activities at different scales may have dependencies or causal relationships, such as taking bread from a bag as part of making a sandwich. 
The system must continuously recognize activities of varying lengths based on newly received video frames.

\begin{figure}[t]
  \centering
  \includegraphics[width=0.49\textwidth]{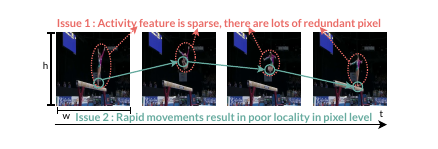}
  \caption{Video frames of the balance beam sports, selected from the FineGym99 dataset~\cite{shao2020finegym}. The entire gymnastic movement was captured in only 17 frames, lasting less than 700 milliseconds.}
  \label{fig:locality}
\end{figure}

\textbf{(ii) Accumulating Redundancy in Streaming Video.} Continuous video activity recognition normally needs to capture meaningful features from the raw streaming videos, which are dominated by noise and redundancy. 
As illustrated in Figure~\ref{fig:locality}, activity-related pixels are sparsely distributed in the video. 
Without effective online reduction, the accumulation of redundant features can quickly overwhelm the activity-related features as the video progresses.

\textbf{(iii) Computational Efficiency Under Resource Constraints.} Continuous video activity recognition must guarantee a timely recognition of the scene while managing the resource constraints of typical edge devices. 
For instance, robots or embodied AI~\cite{duan2022survey} systems must continuously perceive dynamic scenarios through streaming video input to carry out reasoning~\cite{strafforello2023current} and task planning~\cite{gadre2022continuous}.

\begin{figure*}[t]
  \centering
  \includegraphics[width=\textwidth]{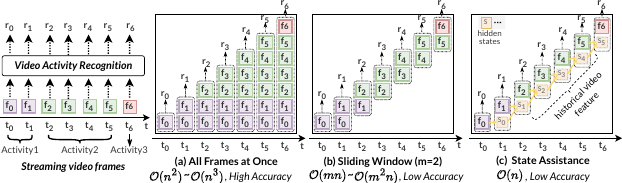}
  \caption{Typical designs for streaming video activity recognition. The ``All Frames at Once'' design processes all video frames at every moment. The ``Sliding Window'' design incorporates only the latest (or sampled) $m$ frames. The ``State Assistance'' relies on a single frame at a time and maintains video context through a hidden state. }
  \label{fig:strawman}
\end{figure*}

Existing video recognition efforts have explored numerous strategies for maintaining video context, yet none have been able to address all the challenges simultaneously.
Works, using ``all frames at once'' strategy (Figure~\ref{fig:strawman}(a)), perform activity recognition on every frame of the video, which theoretically yields the highest accuracy~\cite{slowfast2019fei,kondratyuk2021movinets}. 
However, its efficiency is the biggest concern.
Other works use ``sliding window'' strategy (Figure~\ref{fig:strawman}(b)) to reduce computational cost~\cite{Lin_2019_ICCV,Jiang_2019_ICCV, Li_2020_CVPR, li2022mvitv2}. 
However, these methods struggle to accurately recognize activities at different scales.
Some approaches use a ``hidden state assistance'' design (Figure~\ref{fig:strawman}(c)), where a hidden state compresses video context to enable more efficient activity recognition. 
However, existing state-assisted methods do not effectively eliminate spatiotemporal redundancy in the streaming video, resulting in inefficient utilization of the hidden state and low accuracy.\looseness=-1

\begin{figure}[t]
    \centering
    \includegraphics[width=0.49\textwidth]{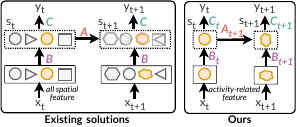}
    \caption{Difference between our idea and existing solutions on maintaining the hidden state.}
    \label{fig:core_idea}
\end{figure}

The all-frame-at-once approach suffers from efficiency degradation over time as data volume grows, while the sliding window method often faces challenges in selecting optimal window sizes or sampling strategies, compromising accuracy. Therefore, our design builds on the existing state-assistant framework, which is efficient—processing one frame at a time—and accurate—adaptively maintaining video context through carefully designed hidden states.
As illustrated in Figure~\ref{fig:core_idea}, we argue that existing state-assistant works lack activity-aware design in both temporal and spatial dimensions. 
They tend to compress all video features, including the redundant background features, into the hidden state, resulting in inefficient use of hidden states, poor preservation of video context, and low recognition accuracy. 
Our core idea is to \textit{selectively compress only activity-related features in temporal and spatial dimensions, which we name adaptive video context modeling}.

In this paper, we propose \sysname for continuous video activity recognition in the streaming video, utilizing the adaptive video context modeling, with two key designs.
\textbf{First, activity spatial feature extraction} (Section~\ref{subsec:stdec}). 
To reduce redundant visual information, we employ a \textit{data reduction strategy that operates without requiring supervision from masks or bounding boxes of activity-related features}. During training, we maximize feature elimination from the video input while preserving recognition accuracy. This adversarial learning process allows our \sysname to discard spatial features irrelevant to the activity.
\textbf{Second, activity-aware state update} (Section~\ref{subsec:ohm}).
Existing methods struggle to effectively maintain video context for multi-scale activities. Our design integrates a \textit{dynamic adaptability, akin to the attention mechanism in transformers}. We dynamically recalculate the weight parameters based on the current input frame during the input processing, hidden state update, and output generation.

We evaluate \sysname across various activity recognition tasks, comparing it with widely-used models for video recognition.
Results show that with significantly fewer trainable parameters (ranging from 8.5k to 34.1M less than the baselines), 
\sysname achieves accuracy improvements of 1.2\% to 79.7\% over baseline methods.
Regarding latency, we evaluate \sysname on typical edge devices (NVIDIA Jetson Orin, Apple MacBook, and an Intel PC with NVIDIA 4090 GPU). It achieves $>$30FPS on all these devices, up to 20$\times$ faster than the available inference systems.

To further explore the capability of our design, we apply \sysname as a video encoder to a large video model to improve its activity recognition capability (Section~\ref{sec:field_study}). Specifically, we replace the visual encoder of the SOTA large video model VideoLlama2~\cite{damonlpsg2024videollama2} by \sysname.  
The results demonstrate a 0.46-point improvement (on a 5-point scale) on the in-distribution dataset, and an improvement ranging from 1.19\% to 4\% on zero-shot datasets. 

\noindent \textbf{Contribution.} We summarize our contribution as follows: 
\begin{itemize}[leftmargin=20pt]
\item We propose a continuous video activity recognition system, \sysname, based on adaptive video context modeling. It promptly recognizes the activities in incoming video frames by efficiently maintaining video context for multi-scale activities.

\item Our \sysname incorporates two key techniques to enhance the utilization efficiency of hidden states, i.e., minimizing redundancy from activity-irrelevant spatial information, and improving adaptability to streaming input while maintaining state.

\item Experiments show that \sysname surpasses all baselines in streaming video activity recognition, using fewer parameters and achieving over 30 FPS on typical edge devices, which can enable on-device video recognition for embodied AI.  We also showcase that  \sysname can enhance the large video models by serving as the video encoder.

\end{itemize}

\section{Background \& Related Works}
\label{sec:background}

Various approaches strive to achieve continuous activity recognition, but none have solved this challenging task.

\begin{figure}[t]
  \centering
  \includegraphics[width=0.46\textwidth]{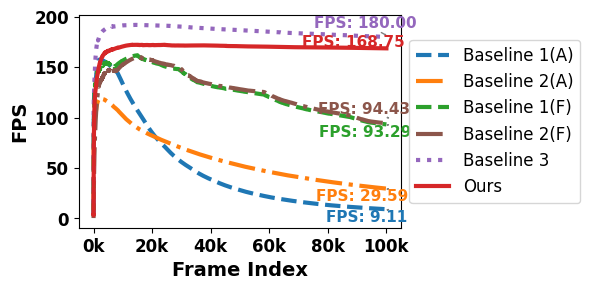}
  \caption{Processing speed of video activity recognition on RTX4090 (a 100k-frame video). Baselines 1(A) and 2(A) represent Transformer and Convpooling-based solutions using the ``All Frames at Once'' design. Baselines 1(F) and 2(F) employ the ``Fixed window size'' design (the window size is 100). Baseline 3 denotes an LSTM-based ``State Assistance'' design.} 
  \label{fig:system_performance_case}
\end{figure}

\noindent \textbf{All Frames at Once.} 
A straightforward method is to input all the past frames into the model at each time step, as illustrated in Figure~\ref{fig:strawman}(a). 
However, it is certainly too costly for model inference and training. Take a Transformer model as an example, with the number of streaming frames $n$ grow, the computation cost of video activity recognition is $O(n^3)$. Even with the many linear alternatives of Transformers, such as Linformer~\cite{wang2020linformer}, RetNet~\cite{sun2023retentive}, Mamba~\cite{gu2023mamba}, and RWKV~\cite{peng2023rwkv}, the cost is still $O(n^2)$. 
Our experiment (Figure~\ref{fig:system_performance_case}) shows that as frames increase, FPS drops to 0.05 times, a 95\% decrease.

\noindent \textbf{Fixed Window Size.} 
The most commonly used technique to control the cost is to fix the input size of the model even with the growing number of frames, such as sliding window or sampling~\cite{Lin_2019_ICCV,Jiang_2019_ICCV, Li_2020_CVPR, li2022mvitv2}, shown in Figure~\ref{fig:strawman}(b). 
For example, TSN~\cite{wang2016temporal} first divides the past video into segments and selects a random snippet from each segment for activity recognition. VideoLlama2~\cite{damonlpsg2024videollama2} only samples 8 or 16 frames as input for any video length. This method has to ignore many detailed features. 
VideoMamba~\cite{li2024videomambastatespacemodel} adopt the Mamba~\cite{gu2023mamba} as the architecture to perform the efficient video understanding, however, it still adheres to the batch-based processing strategy.
\textit{The main challenge is selecting an optimal window size to balance the detection of both short and long activities.}

\noindent \textbf{State Assistance.} 
There are also works~\cite{shi2015convolutional,huang2020efficient,zhou2018temporal}, shown in Figure~\ref{fig:strawman}(c), that only input one current frame into a model (e.g., CNN) first to extract spatial features, 
and then conduct feature aggregation (e.g., through pooling~\cite{kar2017adascan} or LSTM~\cite{hochreiter1997long}) among frames over time to capture temporal dependencies by hidden states. 
By maintaining a group of hidden states, this method significantly enhances efficiency by reducing complexity to approximately $\mathcal{O}(n)$~\cite{shi2015convolutional,huang2020efficient}. However, \textit{as the input size grows, the fixed-size state struggles to remember the increasing history of inputs}~\cite{shao2020finegym,wang2016temporal}.

\section{\sysname Design}


\noindent\textbf{A motivating video sample.}
Video streams contain both spatial (within-frame) and temporal (between-frame) information, both critical for accurate recognition.
Figure~\ref{fig:locality} presents a video sample from a real-world gymnastics match to motivate our design and explain the challenges we faced.
There are two issues we can observe from the video sample. 
\textbf{Issue 1}: the activity-related spatial feature is sparse. The activity feature in a video sample may be captured by only a small portion of the pixel content.
\textbf{Issue 2}: object movement causes poor locality at the pixel level.
It is hard to accurately recognize activities merely by analyzing two adjacent regions (by setting thresholds).

\begin{figure}[t]
  \centering
  \includegraphics[width=0.4\textwidth]{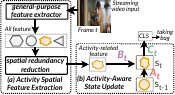}
  \caption{\sysname's workflow for continuous activity recognition via adaptive video context modeling by selectively compressing only activity-related features into the hidden state.}
  \label{fig:system_workflow}
\end{figure}

We have observed the efficiency advantages of state-assistant methods in handling streaming video.
Therefore, we utilize a state-assistant design, which involves recurrent updates over a sequence through hidden states~\cite{sieber2024understanding}. 
Our \sysname has two key designs, i.e., activity spatial feature extraction and activity-aware state update, to solve these two issues. 
Figure~\ref{fig:system_workflow} illustrates the system workflow.

During the usage, streaming video is continuously input into the system frame by frame (without sampling). 
Our \sysname extracts activity-related features and sends them to the activity-aware state update module.
During the state update, the system attempts to identify possible activities and uses the activity features to update the corresponding part of the hidden state. 
If no further activity features related to those features maintained in the hidden state arrive, the corresponding video context is cleared from the hidden state.
The updated state is stored in the system, waiting for the next frame to update it. 
Simultaneously, the updated hidden state can be sent to a classifier for timely activity recognition. 
With our optimization, the entire system can perform real-time video activity recognition on typical edge devices.


\subsection{Activity Spatial Feature Extraction}
\label{subsec:stdec}

The \textbf{challege} of addressing issue 1 is that we \textit{lack supervised data to differentiate between redundant spatial features and those relevant to activities}. 
Because the types of activities in the real world are diverse. 
We cannot merely retain an object's pose and position.
For instance, in Figure~\ref{fig:locality}, it is crucial to also keep the feature of the balance beam present in the environment.
In this way, we can accurately comprehend that this is a gymnastics activity, rather than a simple jump.

The \textit{core idea} of our feature extraction is to utilize the information bottleneck~\cite{tishby2000information} as an \textit{unsupervised method} to extract task-specific features.
Specifically, we employ an information reduction module aimed at minimizing input features while maximizing task accuracy during training. Through this adversarial constraint, it can reduce redundancy and effectively achieve activity-related feature extraction.

Formally speaking, let $X$, $X'$, and $Z$ denote the random variables that represent all spatial features, extracted activity-related features, and redundant features, respectively. 
In our scenario, the activity spatial feature extraction determines how the feature $X'$ is extracted from $X$ for the recognition task $f_\theta$. 
The goal of the activity-related feature extraction is to minimize the inclusion of redundancy information $Z$ in extracted feature $X'$ while ensuring maximum data utility, i.e., 
\begin{equation}
	\min_{\mathcal{I}(X';Z) < \epsilon}\Vert f_\theta(X'), f_\theta(X) \Vert,
	\label{eq:trade-off}
\end{equation}
where $\mathcal{I}$ is the mutual information to quantify the amount of information obtained about one random variable by observing the other random variable. $\Vert \cdot, \cdot \Vert$ is used to measure the distance between two variables.

However, given the lack of supervised data regarding redundant information $Z$, it is hard to minimize the $\mathcal{I}(X';Z)$. 
Note that, $\mathcal{I}(X';Z) \le \mathcal{H}(X')$, where $\mathcal{H}$ is the entropy of a random variable. So, we have the relaxation of Equation~\ref{eq:trade-off}, i.e.,
\begin{equation}
	\min_{\mathcal{H}(X') < \epsilon}\Vert f_\theta(X'), f_\theta(X) \Vert.
	\label{eq:trade-off}
\end{equation}
That is, by minimizing the information of extracted features $X'$ and preserving the accuracy of activity recognition, we can remove the redundant features as much as possible. 
Please note that the data utility objective, $\min\Vert f_\theta(X'), f_\theta(X) \Vert$, is optimized by the accuracy of the activity recognition task (Equation~\ref{eq:loss}).

To reduce the information entropy of extracted features $X'$, we propose two steps, i.e., the pooling layer and an information funnel 
Specifically, 
in our activity spatial feature extraction module, we first a pre-trained powerful feature extractor, i.e., CLIP~\cite{radford2021learning}, to extract general-purpose spatial features from an input frame. 
The CLIP model divides the input frame into multiple patches and computes an embedding for each patch. 
We utilize the average pooling to aggregate these embeddings by computing their mean, generating a single embedding. 
This constitutes the \textit{first} stage of information reduction for $X'$.
After the first stage, each frame is reduced to a single embedding.
Subsequently, this embedding is passed through our information funnel, which consists of a linear layer followed by an activation function, where the output size of the linear layer is smaller than its input size. This completes the \textit{second} stage of information reduction for $X'$.

\subsection{Activity-Aware State Update}
\label{subsec:ohm}

The \textbf{challege} of addressing issue 2 lies in \textit{online associating the features of each incoming frame with those in the video context, i.e., hidden state}, thereby enhancing recognition accuracy.
This state update serves as the association process since the recognition results are based on the updated state.

We take the RNN 
as an example to introduce the traditionalstate update.
It computes $\mathbf{y}$ through a dynamic recurrence of input signals $\mathbf{x}$ at each time step $t$ through hidden states $\mathbf{s}$.
Formally, 
\begin{equation}
\begin{aligned}
    s_t &= \phi_h(\mathbf{A} s_{t-1} + \mathbf{B} x_t + \mathbf{b}_h), \\
    y_t &= \phi_o(\mathbf{C} s_t + \mathbf{b}_o).
\end{aligned}
\label{eq:rnn}
\end{equation}
where $\mathbf{A},\mathbf{B}$, and $\mathbf{C}$ denote the state matrix, input matrix, and output matrix, respectively.
The $\mathbf{b}$ denotes the bias parameters, and the functions $\phi$ denotes activation functions.

However, the RNN and its gating-based variants, e.g., LSTM~\cite{sherstinsky2020fundamentals}, are widely considered challenging to maintain input context by many research works~\cite{greaves2019statistical,zhao2020rnn,miller2018stable}. 
Our experimental results indicate that the LSTM-based baseline performs poorly on streaming video inputs (Table~\ref{tab:long_video}).
We argue that the underperformance of such methods is due to their inefficient utilization of the hidden state. 
Specifically, standard RNN processes input at different moments identically, i.e., fixed $A$, $B$, and $C$ in Equations~\ref{eq:rnn}, leading to irrational use of the hidden state (Figure~\ref{fig:core_idea}). 

We argue that this deficiency stems from the state update method's inability to adapt effectively to incoming video frames, as it applies one update strategy to all frames, irrespective of their distinct features.
We enhance the state update method by enabling activity-aware hidden state updates based on activity spatial features.
The \textit{core idea} is to introduce \textit{dynamic adaptability, akin to the attention mechanism in transformers}, which adjusts its focus across input sequences to accommodate varying inputs.

Specifically, we replace the fixed matrix used for state updates with an input-dependent update manner. 
Our input-dependent update takes the current activity spatial features as input and determines how the current input should be incorporated into the hidden state. Formally, 
\begin{equation}
    A \rightarrow A_t = f_A(x_t); B \rightarrow B_t = f_B(x_t); C \rightarrow C_t = f_C(x_t).
\end{equation}
That is, during training, \textit{we no longer learn the $A$, $B$, and $C$ matrices directly but instead learn how to dynamically generate these matrices based on the input.}

Here, we employ Mamba~\cite{gu2023mamba} as the implementation of our activity-aware state update module, as it aligns with our design philosophy: it dynamically computes these matrices based on each input. 
Additionally, Mamba's implementation accelerates the training process for such recursive formulations.
Note that while Mamba has been explored for video processing~\cite{li2024videomambastatespacemodel,damonlpsg2024videollama2,chen2024video},
these works adopt strategies such as processing ``all frames at once'' or using a ``fixed window size'', treating Mamba primarily as a lightweight alternative to Transformers.
In contrast, our ``state-assistant'' design fundamentally differs in both approach and purpose.

\subsection{Training Structure}
\label{subsec:etd}

The training structure is detailed in Algorithm~\ref{alg:our_train}. 
The \sysname's architecture consists of a pre-trained vision encoder ($\theta_\texttt{CLIP}$),
a feature extractor ($\theta_\texttt{extractor}$), an activity-aware state update module ($\theta_\texttt{state\_maintaince}$), and a classifier ($\theta_\texttt{cls}$).
The vision encoder extracts the general-purpose spatial feature and is frozen during training.
The feature extractor is a linear layer with ReLU activation. 
The activity-aware state update module is a single Mamba layer. 
The classifier is a ReLU activation.

\begin{algorithm}[t]
\small
\DontPrintSemicolon
\KwData{Video frames and corresponding labels: $\texttt{X}_{1:n}$ and $\texttt{Y}_{1:n}$} 
\KwResult{Trained parameters: $\theta$}
\Begin{

\While {not convergent}{
    $\texttt{Emb}_{1:n} \leftarrow \theta_\texttt{CLIP}^\star(\texttt{X}_{1:n})$  \;
    $\texttt{Tok}_{1:n} \leftarrow \theta_\texttt{extractor}(\texttt{Emb}_{1:n})$  \;
    $\texttt{Ir}_{1:n} \leftarrow \theta_\texttt{state\_maintaince}(\texttt{Tok}_{1:n})$  \;
    $\hat{\texttt{Y}}_{1:n} \leftarrow \theta_\texttt{cls}(\texttt{Ir}_{1:n})$ \;
    $\texttt{L} \leftarrow \mathcal{L}(\texttt{Y}_{1:n}, \hat{\texttt{Y}}_{1:n})$ \;
    $\theta \leftarrow \theta - \nabla_\theta \texttt{L}$ \;
}
\KwRet{$\theta$} \label{alg_line:return}
}
\caption{The parallel training structure of \sysname. $\theta_\texttt{spatio\_encoding}^\star$ is frozen during the training.\label{alg:our_train}}
\end{algorithm}

The loss function $\mathcal{L}$ is a frame-weighted loss. 
The core intuition is that activity recognition can be categorized into two types: \textit{motion-focused} (requiring temporal information, e.g., performing a triple somersault) or \textit{scene-focused} (relying solely on static information, e.g., drinking water). 
Formally, it can be expressed as:
\begin{equation}
    \mathcal{L}_v = W^T \times [\mathcal{L}_f^1, \mathcal{L}_f^2, \ldots, \mathcal{L}_f^N],
    \label{eq:loss}
\end{equation}
where $W$ is the weight vector of the loss from Frame 1 to Frame $N$.
For motion-focused tasks, we recommend a linear growth strategy, where the weight of the loss for the initial frame is relatively low and gradually increases over time within the video frames. Specifically, $W = [\frac{i}{N^2}]_{i=1}^N $.
Conversely, for scene-focused datasets, a uniformly weighted approach is recommended. 
That is, $W = [\frac{1}{N}]_{i=1}^N$.

\section{Evaluation}


\noindent \textbf{Hyperparameter Setting.}
The AdamW~\cite{loshchilov2017decoupled} is adopted as the optimizer. The learning rate changes dynamically, using Cosine annealing with warm restarts. This method adjusts the learning rate between 1e-3 and 1e-6 every 10 epochs. 
Aside from these adjustments, the other hyperparameters are kept at their default settings as provided in PyTorch v2.2.0.
Regarding the training process of the network, it requires approximately 100 to 200 epochs to converge in different datasets.
We train these models on a server with 8 RTX4090 GPU cards, and it takes about 3 to 6 hours to train a model on different datasets.

\noindent \textbf{Optimization Deployment.}
We implemented the CLIP vision encoder using the llama.cpp framework~\cite{llama_cpp_impl}, performing INT8 quantization for improved efficiency. 
While the CLIP.cpp framework~\cite{clip_cpp_impl}, based on llama.cpp, only supports CPU backends, limiting GPU use on mobile devices, other frameworks like OpenClip~\cite{open_clip_impl} and ONNXRuntime-GPU~\cite{onnx_impl} also lack support for Jetson GPUs. We extended support for ggml format, enabling GPU-based inference across all platforms, which resulted in a significant performance boost (see Figure~\ref{fig:sys_performance_all}). Furthermore, we compiled the Mamba implementation~\cite{mamba_impl}, which originally supported only PyTorch with CUDA, into ggml format for cross-platform deployment. 
Through these optimizations, we successfully deployed our \sysname on a MacBook Air (M3, 2024), Jetson AGX Orin 64GB, and a PC with an RTX 4090 GPU.

\begin{figure}[t]
  \centering
  \includegraphics[width=0.46\textwidth]{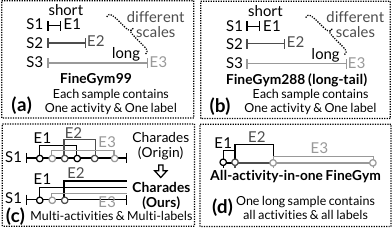}
  \caption{Datasets used in experiments. S represents a sample; a line segment represents the duration of an activity; E represents a label. In Charades, a frame may have multiple labels.}
  \label{fig:dataset}
\end{figure}

\subsection{Experiment Setups}
\label{subsec:expr_setup}

\noindent \textbf{Evaluation Goals.}
We aim to evaluate \textbf{four capabilities}:

\noindent\textit{Capability 1}: Recognition of activities across timescales. \looseness=-1

\noindent\textit{Capability 2}: Activity Recognition in streaming input.

\noindent\textit{Capability 3}: Prompt response as more frames are received.

\noindent\textit{Capability 4}: Generality for long-tail activities.

\noindent\textbf{Datasets.}
We utilize five datasets to evaluate different capabilities. 
\textit{FineGym99}~\cite{shao2020finegym} (For capabilities 1 \& 3). It is a gymnasium video dataset. 
Each video clip is labeled with a gymnastics movement. 
The duration of activities varies from 2 to 50 seconds. 
\textit{FineGym288}~\cite{shao2020finegym} (For capabilities 1, 3, \& 4). It is an extended version of FineGym99, which enhances its by adding categories. 
FineGym288 has a natural long-tail distribution.

\textit{Charades}~\cite{sigurdsson2016hollywood} (For capabilities 1 \& 3). It is a multi-label dataset, composed of daily indoor activities with an average length of 30 seconds. 
Each video clip consists of about 800 frames on average, varying from 32 to 3,944 frames. 
We re-annotated Charades, as Figure~\ref{fig:dataset}(b), to evaluate the method's effectiveness in maintaining video context when multiple activities occur simultaneously.

\textit{All-activity-in-one FineGym} 
(For capabilities 1, 2, \& 3). It is constructed based on the FineGym99 as shown in Figure~\ref{fig:dataset}(c). 
We concatenated all the video clips from FineGym99's testing dataset, forming a single long video lasting 3.5 hours and consisting of over 440k frames. 
The concatenated video is employed to simulate streaming video containing continuous activities.

\begin{figure}[t]
    \centering
    \includegraphics[width=0.42\textwidth]{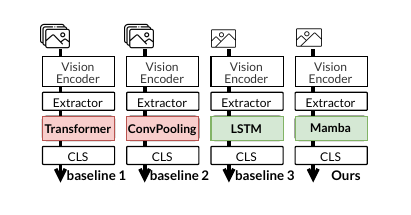}
    \caption{The constructed baselines 1, 2, and 3.}
    \label{fig:baselines}
\end{figure}

\noindent\textbf{Baseline Methods.}
We utilize five baselines to execute continuous perception, whereby the method predicts an outcome at each timestep. Baselines 1, 2, and 3 are constructed following different video processing paradigms (as shown in Figure~\ref{fig:baselines}).
Their overall architecture \textit{mirrors ours}, except that we replace the mamba layer with other networks. 
\textit{Baseline 1} utilizes a Transformer~\cite{vaswani2017attention}. \textit{Baseline 2} utilizes a convPooling~\cite{kar2017adascan}, which consists of a 1D convolution and a 1D adaptive max pooling. \textit{Baseline 3} utilizes an LSTM~\cite{hochreiter1997long}.

Baselines 4 and 5 are state-of-the-art (SOTA) methods for balancing accuracy and efficiency for video classification for on-device computing scenarios. 
\textit{Baseline 4:} MoviNetA0 \cite{kondratyuk2021movinets}. This is a Mobile Video Network specifically designed for efficient video recognition.
This method applies down-sampling when processing videos. Drawing from the experience of using this model on the Charades dataset, we similarly down-sample the video to 6 fps for video preprocessing.
\textit{Baseline 5:} Mvitv2\cite{li2022mvitv2} This is an improved multiscale vision transformer for both classification and detection tasks.
This method necessitates the sampling of any given video to a fixed number of frames for video classification tasks. We trained the baseline that takes as input 16 frames at a time, which is one of the default settings.

\noindent\textbf{Input Strategy.}
We utilize three typical input strategies to perform the evaluation. 

\textit{S1: All frames feeding.}
All frames from frame 1 up to frame $N$ are fed into the baseline method to provide the necessary context for achieving continuous perception at timestep $N$.
\textit{S2: Sliding windowed feeding.}
All frames between the $(N-m)$-th frame to the $N$-th frame are fed into the baseline at timestep $N$, where $m$ denotes the sliding window size. 
\textit{S3: Single frame feeding.}
Only frame $N$ is supplied to the baseline method.

\noindent\textbf{Metrics.}
In this work, we utilize various metrics, namely Accuracy, Mean Average Precision (mAP), and the Jaccard Index for single-label and multi-label video classification tasks. 
Furthermore, we introduce a novel metric referred to as the \textit{Early Detection Rate (EDR)}, designed to evaluate the model's ability to detect activities promptly within videos. 
The definition is $\mathtt{EDR} = {(E_{pred} - E_{gt})}/{E_N}$,
where $E_{pred}$ refers to the frame index at which the label is detected. 
$E_{gt}$ represents the ground truth frame index, that is, the index of the frame where the activity begins. 
$E_N$ signifies the total number of frames in the activity.


\begin{table*}[t]
\centering
\caption{The accuracy and EDR results of different baselines. The \#Param counts the parameters that need to be trained. Baselines 1, 2, 3, and our method have the same pre-trained clip with a parameter size of 149M.}
\label{tab:main_acc}
\small
\begin{tabular}{l|l|ll|ll|lll}
\hline
\multicolumn{1}{c|}{{ }}                            & \multicolumn{1}{c|}{{ }}                                   & \multicolumn{2}{c|}{{ \textbf{FineGym99}}}                                 & \multicolumn{2}{c|}{{ \textbf{FineGym288}}}                                & \multicolumn{3}{c}{{ \textbf{Charades}}}                                                                                                         \\ \cline{3-9} 
\multicolumn{1}{c|}{\multirow{-2}{*}{{ \textbf{Methods}}}} & \multicolumn{1}{c|}{\multirow{-2}{*}{{ \textbf{\#Param}}}} & \multicolumn{1}{l|}{{ \textbf{Acc (\%)}}} & { \textbf{EDR (\%)}} & \multicolumn{1}{l|}{{ \textbf{Acc (\%)}}} & { \textbf{EDR (\%)}} & \multicolumn{1}{l|}{{ \textbf{$\text{mAP}_\text{all}$ (\%)}}} & \multicolumn{1}{l|}{{ \textbf{$\text{mAP}_\text{last}$ (\%)}}} & { \textbf{EDR} (\%)} \\ \hline
{\small{\textbf{B1} (Transformer-based)}}                              & { 
\textbf{210.5K}}                                           & \multicolumn{1}{l|}{{ 82.3}}         & { 48.20}      & \multicolumn{1}{l|}{{ 76.6}}         & { 53.20}      & \multicolumn{1}{l|}{{ 17.2}}              & \multicolumn{1}{l|}{{ 21.4}}               & { 52}       \\ \hline
{\small{\textbf{B2} (ConvPooling-based)}}                              & { \textbf{367.3K}}                                           & \multicolumn{1}{l|}{{ 82.6}}         & { 38.40}      & \multicolumn{1}{l|}{{ 75.4}}         & { 41.40}      & \multicolumn{1}{l|}{{ 15.8}}              & \multicolumn{1}{l|}{{ 21}}                 & { 43.3}       \\ \hline
{\small{\textbf{B3} (LSTM-based)}}                              & { \textbf{218K}}                                             & \multicolumn{1}{l|}{{ 84.1}}         & { 23.40}      & \multicolumn{1}{l|}{{ 78.4}}         & { 25.80}      & \multicolumn{1}{l|}{{ 15.33}}             & \multicolumn{1}{l|}{{ 17.8}}               & { 84}         \\ \hline
{\small{\textbf{B4} (MoviNet)}}                              & \textbf{ 2.1M}                                                    & \multicolumn{1}{l|}{{ 19.7}}         & { 41.60}      & \multicolumn{1}{l|}{{ 13.9}}         & { 44.20}      & \multicolumn{1}{l|}{{ 3.9}}               & \multicolumn{1}{l|}{{ 5.8}}                & { \textbf{32}}         \\ \hline
{\small{\textbf{B5} (Mvitv2)}}                              & \textbf{ 34.3M}                                                   & \multicolumn{1}{l|}{{ 6.4}}          & { \textbf{20.50}}      & \multicolumn{1}{l|}{{ 3.1}}          & { \textbf{24.10}}      & \multicolumn{1}{l|}{{ 5.3}}               & \multicolumn{1}{l|}{{ 7.3}}                & { 99.9}       \\ \hline
{ \textbf{Ours}}                                    & { \textbf{202K}}                                             & \multicolumn{1}{l|}{{ \textbf{86.1}}}         & { 22.50}      & \multicolumn{1}{l|}{{ \textbf{81.3}}}         & { 25.10}      & \multicolumn{1}{l|}{{ \textbf{18.4}}}              & \multicolumn{1}{l|}{{ \textbf{22.1}}}               & { 56.3}       \\ \hline
\end{tabular}
\vspace{-10pt}
\end{table*}

\subsection{Evaluation of All Capabilities} 
\label{subsec:main_res}

In Table~\ref{tab:main_acc}, we present the outcomes from several baseline models across different datasets. We first introduce each column's meaning and statistical methods, followed by an evaluation of our methodology and findings.

The \#Param column signifies the size of a model's parameters. Owing to the varying category numbers in the three datasets, there is a marginal difference in the parameter count in the final layer for different tasks, approximately a 10k difference in parameters.
Here, we reference the parameter count when classifying the FineGym99 dataset. Baselines 1, 2, and 3, as well as our model, all utilize the same CLIP model with a parameter size of 149M.
When excluding the CLIP, our design exhibits the smallest parameter count.

We chose the most natural input strategy for each baseline to assess the accuracy. For batch-based methods, i.e., Baselines 1, 2, 4, and 5, we adopted the all-frame feeding strategy. Although this significantly increases computational overhead, it enables these methods to have sufficient context to recognize the video activities. Both Baseline 3 and our method utilize a streaming processing paradigm, for which we selected the single-frame feeding strategy.

In evaluating single-label classification datasets, we utilize Accuracy (Acc) as the primary measure of correctness. This accuracy is ascertained by verifying whether the correct action is detected within a given video. 
The EDR is calculated as introduced in Section~\ref{subsec:expr_setup}(Metrics). 
When dealing with multi-label classification datasets, two variants of mean Average Precision (mAP) are employed to gauge correctness.
The $\text{mAP}_\text{all}$ metric is determined by considering predictions of all frames in the video. 
The $\text{mAP}_\text{last}$ entails calculations based solely on the prediction results of the final frame in the video. It's important to note that the label of the last frame includes labels for all actions present within the current video.
Each label's EDR is computed separately and then averaged. 
If a method fails to detect a particular label, this failure is taken into account during computation by adding a 100\% increase to the result for that category. This signifies that the correct label was undetected even after the video's conclusion.
Below we present our experimental analyses and the subsequent conclusions:

\textbf{Superior Accuracy:} Our method consistently outperforms all baseline methods in terms of accuracy across all tasks, improving metrics by ranges of 2 to 79.7, 2.9 to 78.2, 1.2 to 14.5, and 0.7 to 16.3. The leading accuracy demonstrates that our \sysname can effectively recognize activities of different scales (\textit{Capability 1}).
The superior accuracy on the FineGym288 dataset underscores our method's generalization ability for long-tailed data (\textit{Capability 4})).

\textbf{Prompt Recognition :} Our approach exhibits an outstanding ability to promptly detect activities (\textit{Capability 3}) and continuously process (\textit{Capability 2}) by EDR. For single-label tasks, activities are detected within the first quarter of their occurrence. For complex multi-label tasks, all activities are identified halfway through the video.

\textbf{Optimal Use of Trainable Parameters:} Notably, our method retains superior performance while utilizing a minimal number of training parameters, amounting to only 202K. This represents a drastic reduction when compared to the parameter requirements of baseline models, with differences ranging from 8.5K to 34.1M.

\textbf{Robust Stability Across Varied Tasks:} Our method showcases robust stability in tackling diverse tasks. It performs uniformly well across differing task scales (including those with varied activity durations), varied video characteristics (motion-focused and scene-focused video), and differing classification objectives (such as single-label and multi-label classifications). No baseline methods are currently able to match this level of comprehensive performance.

\textbf{Maintaining Video Context for Multiple Activities:} 
Within the multi-label dataset Charades, the $\text{mAP}_\text{last}$ denotes the ability to recognize all activities in the video by its conclusion. Notably, Baselines 1 and 2 attained such accuracy by taking as input all video frames simultaneously. Our method achieves a higher accuracy based on the last frame and implicit memorization of past frames. Notably, our approach achieves 1.24 times the accuracy of Baseline 3, which employs LSTM for continuous processing. Results show that our \sysname shows a better utilization of the hidden state. 

Considering the inadequate accuracy demonstrated by Baselines 4 and 5, subsequent experimental evaluations will only involve Baselines 1, 2, and 3, along with ours.

\begin{table}[t]
\small
\caption{Evaluation results on All-activity-in-one FineGym. (The parenthesized number of S2 signifies the window size, while 725 refers to the maximum frame count of a single activity.)\label{tab:long_video}}
\begin{tabular}{l|lllll|l}
\hline
\textbf{Method}                                                                        & \multicolumn{5}{c|}{\textbf{Baseline 1}}                                                                                                                                                                                                                                                                                      & \textbf{Baseline 3} \\ \hline
\multicolumn{1}{c|}{\textbf{\begin{tabular}[c]{@{}c@{}}Input\\ Strategy\end{tabular}}} & \multicolumn{1}{l|}{\textbf{S3}} & \multicolumn{1}{l|}{\textbf{\begin{tabular}[c]{@{}l@{}}S2 \\ (10)\end{tabular}}} & \multicolumn{1}{l|}{\textbf{\begin{tabular}[c]{@{}l@{}}S2 \\ (100)\end{tabular}}} & \multicolumn{1}{l|}{\textbf{\begin{tabular}[c]{@{}l@{}}S2 \\ (725)\end{tabular}}} & \textbf{S1}                     & \textbf{S3}         \\ \hline
\textbf{Acc (\%)}                                                                      & \multicolumn{1}{l|}{4.1}         & \multicolumn{1}{l|}{6.6}                                                         & \multicolumn{1}{l|}{4.2}                                                          & \multicolumn{1}{l|}{6.2}                                                          & \multicolumn{1}{c|}{\textbf{-}} & 7                   \\ \hline
\textbf{Method}                                                                        & \multicolumn{5}{c|}{\textbf{Baseline 2}}                                                                                                                                                                                                                                                                                      & \textbf{Ours}       \\ \hline
\multicolumn{1}{c|}{\textbf{\begin{tabular}[c]{@{}c@{}}Input\\ Strategy\end{tabular}}} & \multicolumn{1}{l|}{\textbf{S3}} & \multicolumn{1}{l|}{\textbf{\begin{tabular}[c]{@{}l@{}}S2 \\ (10)\end{tabular}}} & \multicolumn{1}{l|}{\textbf{\begin{tabular}[c]{@{}l@{}}S2\\ (100)\end{tabular}}}  & \multicolumn{1}{l|}{\textbf{\begin{tabular}[c]{@{}l@{}}S2\\ (725)\end{tabular}}}  & \textbf{S1}                     & \textbf{S3}         \\ \hline
\textbf{Acc (\%)}                                                                      & \multicolumn{1}{l|}{4.7}         & \multicolumn{1}{l|}{11.8}                                                        & \multicolumn{1}{l|}{13.4}                                                         & \multicolumn{1}{l|}{11.3}                                                         & 7                               & \textbf{14.8}       \\ \hline
\end{tabular}
\end{table}

\begin{figure*}[t]
  \centering
  \includegraphics[width=\textwidth]{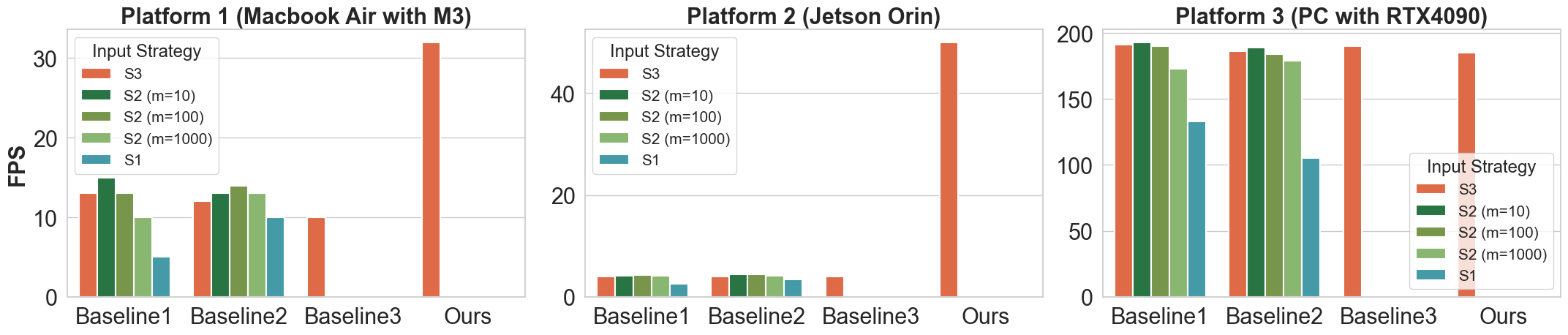}
  \caption{Processing speed of streaming video activity recognition with 10k frames on different platforms.}
  \label{fig:sys_performance_all}
\end{figure*}

\subsection{Further Evaluation of Capabilities 1 \& 2}
\label{subsec:continous_perception}
We assess the continuous perception capability with the All-activity-in-one FineGym dataset, a 3.5-hour video described in Section~\ref{subsec:expr_setup}.
All baseline methods use the model trained on the FineGym99 dataset, whose accuracy evaluated on a single video clip is reported in Table~\ref{tab:long_video}.
These methods are evaluated with different input strategies. 
For batch processing approaches (Baselines 1 and 2), we test all three strategies. 
Regarding streaming processing methods (our \sysname and Baseline 3), we use the single frame feeding strategy (Strategy 3).
Both our \sysname and Baseline 3 benefit from the model's memorization of past video frames.
During continuous perception, each frame is processed only once without redundant input.
The Acc metric is calculated by considering predictions of all video frames. 

Baseline 1's transformer structure results in slower processing as the input sequence grows.
Therefore, Baseline 1 faced challenges during evaluation with input strategy 1 due to increasing input sequence length over time.
Detailed system performance is provided in Figure~\ref{fig:system_performance_case}. 

Based on the results, we draw the following two conclusions. 
First, our \sysname demonstrates the \textbf{best continuous perception ability} (\textit{Capability 2}). It surpasses batch processing methods ranging from 0.5 to 10.7. Its accuracy is twice that of traditional LSTM-based baseline methods. 
Second, our approach can effectively \textbf{adapt to activities of varying scales} (\textit{Capability 1}). In the hours-long video, activities lasted from tens of frames to hundreds of frames. Our \sysname achieves the best accuracy, which outperforms the current sliding-window and state-assist methods.

\subsection{Real-time Inference on Edge}
\label{sec:performance}

We measure the system performance on three mobile platforms with two long video clips. 
The long video clips were excerpted from the All-activity-in-one FineGym dataset, one containing 10k frames and the other 100k frames. 
We conducted each experiment three times and calculated the average.
In Figure~\ref{fig:sys_performance_all}, we present the system performance of different baselines processing 10k frame videos on the three platforms. 
Additionally, in Figure~\ref{fig:system_performance_case}, we report the system performance of different baselines processing 100k frame videos on a PC equipped with an RTX4090. 
FPS is calculated by dividing the total number of video frames by the total processing time.

Our experimental findings are as follows.

\textbf{Recognizing the video stream in real time.} In Figure~\ref{fig:sys_performance_all}, it can be observed that our method achieves real-time processing ($>$30fps) on different platforms. Even on Macbook Air and Jetson, our speed is 2 to 20 times faster than the baseline methods. This improvement stems from two aspects. \textit{Firstly, our streaming processing method} eliminates the need to use the model through all feeding or sliding window approaches. \textit{Secondly}, in terms of \textit{system implementation and optimization}, recall that mainstream frameworks, e.g., OpenClip, clip.cpp, and ONNXRuntime-GPU, lack support for using the CLIP model with Jetson GPU. Therefore, baselines can only utilize the CPU for computation. Our prototype can enable GPU computation on both Macbook Air and Jetson. \looseness=-1

\textbf{No significant performance degradation as the frame number increases.} As shown in  Figure~\ref{fig:system_performance_case}, with an increase in input sequences, the performance of Baselines 1 and 2, using the all frame feeding and sliding windowed feeding strategy, significantly declined. When processing a 100k frame video, Baseline 1 drops to a speed of 9fps.

\section{Case Study: Video Encoder for Large Video Models}
\label{sec:field_study}
\begin{figure}[t]
  \centering
  \includegraphics[width=0.49\textwidth]{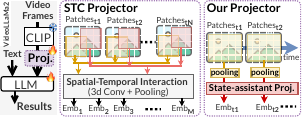}
  \caption{Applying our \sysname into VideoLLaMa2 as a video encoder.}
  \label{fig:video_llama}
  
\end{figure}

In this section, we discuss how the design of our system, \sysname, improves large video models in video activity recognition and understanding. 
As a case study, we analyze one of the SOTA large video models, VideoLLaMa2~\cite{damonlpsg2024videollama2} (left side of Figure~\ref{fig:video_llama}). 
During video processing, VideoLLaMa2 first samples $N$ frames from the video. These frames are then converted into $M$ tokens by a multimodal projector (denoted as "Proj." in the figure). Simultaneously, the user's text prompt is tokenized, combined with the video tokens, and input into the LLM for inference. 


\subsection{Problem Analysis}

VideoLLaMa2 introduces an STC projector as the projection layer. 
As shown in the middle part of Figure~\ref{fig:video_llama}, the STC projector utilizes a \textit{batch-based idea} to understand the video \textit{without reducing the redundancy}. 
Specifically, it performs 3D convolutions on the feature of $N$ sampled frames.
The design of \textbf{the STC projector limits VideoLLaMa2's effectiveness and performance}.
Therefore, it struggles to support higher sampling rates because 3D convolutions on a larger batch demand significant computational resources. As a result, only two versions of the model are available: one that processes 8 frames and another that processes 16 frames. Regardless of the video length or event scale, this fixed frame sampling inevitably reduces the model's accuracy.

\subsection{Our design}
We adopt the design of our \sysname to build a new projection layer, named the \sysname projector,
which replaces the STC projector. 
This new projection utilizes our activity-aware state update to maintain the video context efficiently.
Our state-assisted design can efficiently increase the sampling rate of the input video, far exceeding the existing designs with 8 or 16 frames, without increasing the processing time.
Figure~\ref{fig:video_llama} (right) illustrates our design. 
When a video frame is input, the projector first processes the spatial information within the frame by applying mean pooling to patch embeddings. 
Next, our state-assistant module processes the input video, capturing relationships over time. 

\begin{table}[t]
\centering
\caption{Evaluation results on the VideoLLaMa2.}
\label{tab:case_study}
\begin{tabular}{l|l|l|l}
\hline
                     & \small{\begin{tabular}[c]{@{}l@{}}\textbf{VideoChatGPT} \\ Corr. (5-point)\end{tabular}} & \small{\begin{tabular}[c]{@{}l@{}}\textbf{ActivityNet} \\ Acc (\%)\end{tabular}} & \small{\begin{tabular}[c]{@{}l@{}}\textbf{MSVD} \\ Acc (\%)\end{tabular}} \\ \hline
\textbf{\small{VideoLLaMa2}} & 1.95                                                                             & 38.46                                                                 & 44.6                                                             \\ \hline
\textbf{\small{Ours}}        & \textbf{2.41}                                                                    & \textbf{39.65}                                                          & \textbf{48.6}                                                    \\ \hline
\end{tabular}
\end{table}

\subsection{Experiments}

\noindent \textbf{Implementation.}
Using the official videollama2 codebase~\cite{video_llama_impl}, we implemented our \sysname projector.
This projector consists of mean pooling, a linear layer with ReLU, and the history modeling layer of our \sysname with a code dimension of 2048, followed by another linear layer with ReLU. The total parameter count of our projector is 450\,M. For comparison, the STC projector with 8 frames has 1.8\,B parameters, which is four times larger.

We fine-tuned and evaluated the model following the official guidelines. All hyperparameters were set to the official default values. During fine-tuning, only the large language model (LLM) and the projection layer were trained. We used the videoChatGPT dataset~\cite{maaz2023video} for fine-tuning both the original model structure and our proposed model structure. 
All experiments were performed on a machine with eight A600 GPUs. Due to limited training resources, we used the 8-frame model as the baseline, while our method extended the training sample length to 128 frames.

\noindent \textbf{Results.}
Testing was conducted on the videoChatGPT dataset (in-distribution), ActivityNet-QA~\cite{yu2019activityqa} (zero-shot), and MSVD-QA~\cite{xu2017video} (zero-shot), and the results are reported in Table~\ref{tab:case_study}. 
For the VideoChatGPT dataset, correctness was measured on a 5-point scale. Our method outperformed the baseline by 0.46 points. 
On the other two zero-shot datasets, our approach achieved improvements of 1.19\% and 4\%, respectively. 
These results demonstrate that our projector design effectively improves model accuracy, highlighting a new design concept for video projection for large video models.

\section{Conclusion}



Continuous video activity recognition in streaming video is important for complex multimodal human-computer interaction, e.g., embodied AI.
However, achieving this on resource-limited edge devices is challenging. 
This paper proposes \sysname, a state-assistant approach for continuous video activity recognition in streaming video.
It consists of two key designs: activity spatial feature extraction to reduce spatial redundancy and activity-aware state update to maintain the video context effectively.
Experimental results show that \sysname can achieve real-time continuous video recognition on typical edge platforms. 
We believe that our \sysname offers a novel idea for future streaming video processing.

{
    \small
    \bibliographystyle{ieeenat_fullname}
    \bibliography{main}
}

\end{document}